\documentclass{article}

\usepackage[nonatbib, final]{neurips_data_2024}
\usepackage[numbers]{natbib}

\usepackage[ruled,linesnumbered]{algorithm2e}

\usepackage[utf8]{inputenc} 
\usepackage[T1]{fontenc}    
\usepackage{url}            
\usepackage{booktabs}       
\usepackage{amsfonts}       
\usepackage{nicefrac}       
\usepackage{microtype}      
\usepackage[dvipsnames]{xcolor}
\usepackage{enumitem}
\usepackage{graphicx}
\usepackage{xspace}
\usepackage{soul}
\usepackage{multirow}
\usepackage{bbold}
\usepackage{makecell}
\usepackage{hyperref}       
\usepackage{amsmath}
\usepackage{cleveref}
\usepackage{bm}

\SetKwRepeat{Do}{do}{while}%

\title{Digital Player: Evaluating Large Language Models based Human-like Agent in Games}

\author{%
  $\,$ Jiawei Wang$^{1,2}$,Kai Wang$^1$, Shaojie Lin$^1$, Runze Wu$^1$,  Bihan Xu$^1$, Lingeng Jiang$^1$, Shiwei Zhao$^1$, \\ \bf Renyu Zhu$^1$,  Haoyu Liu$^1$, Zhipeng Hu$^1$, Zhong Fan$^1$, Le Li$^1$, Tangjie Lyu$^1$, Changjie Fan$^1$ \\
  $^1$Fuxi AI Lab, NetEase Games \\ $^2$Hangzhou Institute for Advanced Study, University of Chinese Academy of Sciences\\
  \texttt{\small \{wangjiawei14,wangkai02,linshaojie01,wurunze1,xubihan,jianglingeng,zhaoshiwei,}\\
  \texttt{\small zhurenyu,liuhaoyu03,zphu,fzn0710,lile,hzlvtangjie,fanchangjie\}@corp.netease.com} \\ 
}
\begin{document}

\maketitle

\begin{abstract}
With the rapid advancement of Large Language Models (LLMs), LLM-based autonomous agents have shown the potential to function as digital employees, such as digital analysts, teachers, and programmers. In this paper, we develop an application-level testbed based on the open-source strategy game ``Unciv'', which has millions of active players, to enable researchers to build a ``data flywheel'' for studying human-like agents in the ``digital players'' task.
This ``Civilization''-like game features expansive decision-making spaces along with rich linguistic interactions such as diplomatic negotiations and acts of deception, posing significant challenges for LLM-based agents in terms of numerical reasoning and long-term planning. Another challenge for ``digital players'' is to generate human-like responses for social interaction, collaboration, and negotiation with human players.
The open-source project can be found at \textbf{ \textit {\url{https:/github.com/fuxiAIlab/CivAgent}}}.
\end{abstract}

\section{Introduction}
\label{sec1}
With the advent of ChatGPT~\cite{chatgpt} and GPT-4~\cite{achiam2023gpt}, large language models (LLMs) have made remarkable strides, showcasing their potential to achieve human-like intelligence~\cite{chang2024survey}. A variety of benchmarks have been proposed to compare different LLMs and to investigate the limits of this approach, ranging from natural language processing (NLP) tasks~\cite{hendrycks2020measuring, zhang2024m3exam} to mathematics~\cite{hendrycks2021measuring,mishra2022lila}. In contrast to focusing on LLMs' capacity for general problems, there has been a growing research area that employs LLMs as central controllers to construct autonomous agents for challenging specific tasks within interactive environments with the goal of obtaining human-like decision-making capabilities~\cite{park2023generative, wang2023voyager, agent2024survey}. 
The key idea is to equip LLMs with crucial human capabilities like memory~\cite{park2023generative}, reflection~\cite{shinn2023reflexion}, and planning~\cite{yao2022react} under certain profilings.  There is also a growing recognition of the importance of agent architectures beyond LLMs, such as Retrieval-Augmented Generation (RAG)~\cite{gao2023rag}, workflows~\cite{zeng2023flowmind}, multi-agent collaboration~\cite{hong2023metagpt, wu2023autogen}, and LLMs ensemble~\cite{jiang2023llm}. For instance, it has been observed that state-of-the-art LLM-based agents~\cite{zhou2023language, shinn2023reflexion} for code generation achieved scores exceeding 90 on the HumanEval benchmark~\cite{chen2021humaneval}, significantly outperforming the vanilla ChatGPT (50 points) and GPT-4 (67 points). 
LLM-based agents are considered to be one of the key technologies for achieving Artificial General Intelligence (AGI), and one of its exciting prospect is their application across various industries as domain-specific human-like proxies~\cite{qian2023communicative, li2024doctor, huang2023agentcoder}, commonly referred to as ``digital employees''. The agents are not only expected to possess expertise in specific fields but also exhibit anthropomorphic qualities for human-AI interaction. We refer to these as LLM-based human-like agents, distinguishing them from more generic LLM-based agents.

Inherited the data-centric research methodology of LLMs, for human-like agent development, human feedback data is considered crucial for addressing the limitations of current LLMs. The industrial consensus posits that the gold standard for agent development is to create a ``data flywheel''~\cite{mahendra2023building, achiam2023gpt} where real human feedback data such as conversational logs from chatbots is utilized to enhance agents through methods like fine-tuning~\cite{hu2023llm}, RAG, and workflow optimizations, and improved agents then attracts more users. Contrary to relying on static datasets, data flywheels emphasize iterative improvements driven by human-AI interactions. However, due to its commercial value, there is a scarcity of existing work provide access to substantial data.  And owing to the cost of collecting human feedback data, it is hard for non-commercial researchers to establish a data flywheel.

However, unbeknownst to many,
``citizen science'' games~\cite{good2011games,sullivan2018deep,waldispuhl2020leveling} or ``crowdsourcing games'' have, in recent years, successfully served as a cost-free mechanism for large-scale crowdsourced contributions to various scientific research endeavors. An example is seen in Foldit, where 57,000 players produced valuable results that either matched or surpassed algorithmically computed solutions~\cite{cooper2010predicting}. Additionally, EteRNA~\cite{treuille2014scientific} has attracted over 26,000 participants who have contributed RNA sequences that conform to a specified structure. 
These facts suggest that gaming has the potential to assist researchers without commercial backing in constructing low-cost data flywheels for tasks involving LLM-based human-like agents.

In fact, games have long been considered an excellent testbed for artificial intelligence (AI) research due to their harmless nature and scalability, particularly in the realm of reinforcement learning (RL)~\cite{vinyals2017starcraft, meta2022diplomacy,dqn}. In the LLM-based agents domain, text-based games~\cite{light2023avalonbench,xu2023exploring,wang2023avalon} are used to create text environments for language agents to chat with human players. Voyager~\cite{wang2023voyager} constructs a self-improving evolutionary agent in the popular game Minecraft, but with no involvement of humans. The most analogous work to this paper is Generative Agents~\cite{park2023generative}, which seeks to simulate a small town populated by approximately a dozen human-like agents, observing whether phenomena similar to those found in human societies might emerge. However, given the current advancements in LLM-based agents, the town environment no longer presents sufficient challenges. Its primary shortcomings include: 1) Lack of quantitative assessment. Phenomena such as information dissemination within the town require manual crowdsourcing for evaluation; 2) Absence of a numerical system. Agents in the town lack explicit objectives and thus there is no quantifiable metrics to assess their reasoning and planning capabilities; 3) Unengaging. Although initially generating buzz, the simulated society falls short of being engaging, making it difficult to attract players for human feedback data generation. 
\begin{figure*}[tb]
  \centering
  \includegraphics[scale=0.16]{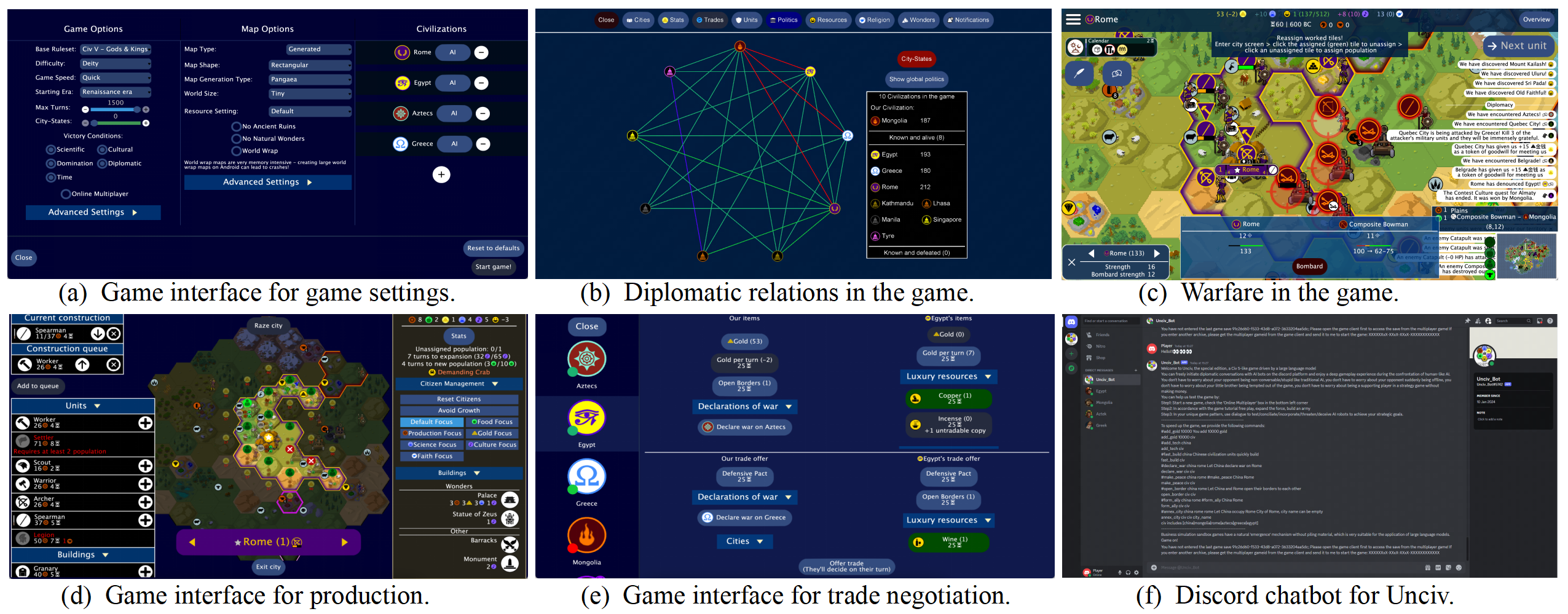}
  \caption{Examples of interfaces of Unciv game and Discord chatbots.\label{Fig:game}}
\end{figure*}

In this paper, we focus on the instantiation of the digital employee concept within gaming environments by developing a LLM-based human-like agent that operates as a ``digital player''.
We introduce CivSim, which is built upon the widely played strategy game ``Unciv''\footnote{https://github.com/yairm210/Unciv}, boasting millions of active players. Unciv is an open-source homage to the renowned ``Civilization'' series, where each agent takes on the role of leading a civilization from the ancient age to the information age (Figure \ref{Fig:game}(a)). The game's rules closely mirror those mechanisms that govern human societies: civilizations must strike a balance between economic development (Figure \ref{Fig:game}(d)), scientific research, diplomacy (Figure \ref{Fig:game}(b)), and warfare. Players are faced with decisions ranging from long-term strategic planning such as diplomatic negotiations (Figure \ref{Fig:game}(e)) to detailed tactical moves like controlling individual units (Figure \ref{Fig:game}(c)). As a turn-based strategy game, Unciv emulates the progression of human history and societal evolution over hundreds of turns that can span several hours to days, making it challenging  for players to find friends to engage in multiplayer battles. This extended duration also underscores the necessity for profound reasoning and extensive strategic foresight. In addition to significantly reducing agent numbers, thereby lowering costs and enhancing economic viability, CivSim sets itself apart with distinctive features:

\begin{itemize}[leftmargin=0pt]
    \item Language-based diplomatic negotiations and deception. The game provides elements for negotiation. Successful negotiation involves identifying the opponent's bottom line and expanding one's profit margin through negotiation skills~\cite{abdelnabi2023llm, kwon2024llms}. Deception involves injecting false information into the agent's memory through dialogue, potentially leading to decision changes if the opponent fails to recognize the false information.
    \item Numerical reasoning and long-term planning. In Unciv, the decisions space of players covers production, technology and culture development, diplomatic negotiations, and warfare, all of which are extensive. Players' decision-making relies on the gaming context and self-information, both represented in multidimensional numerical forms. However, research has revealed certain shortcomings in LLM's reasoning capabilities on numerical questions~\cite{valmeekam2022large, huang2022towards}. Long-term planning presents another challenge within the game due to the inherent strategic nature and dynamic diplomatic relations within game. Players must navigate between being a reliable ally and reneging on commitments --- a decision that carries long-term consequences. Indeed, complex strategy game remains a challenge, even for reinforcement learning algorithms~\cite{meta2022diplomacy,qi2023civrealm, gomme2024player}.
    \item Anthropomorphism. Differs from simulations purely composed of LLM-based agents, we aim to create ``digital players'' that require human involvement. Unlike language-based Turing Tests~\cite{jones2023does}, we aim for creating agents with human-like behavior under complex scenarios, including cooperation and competition with humans~\cite{xu2023exploring,light2023avalonbench}. Due to human player involvement, the agent's anthropomorphic abilities will impact the outcome of the game through diplomatic relations between players. 
\end{itemize}

To foster future research on Unciv, we introduce CivAgent, a human-like digital player designed for turn-based strategy games. 
We develop foundational capabilities for CivAgent to engage players and initiate the data flywheel. 
Our implementation utilizes OpenAI's ChatGPT as the default foundational LLM, LlamaIndex\footnote{https://www.llamaindex.ai/} for memory, RAG and expert workflows within the agent, and Ollama\footnote{https://ollama.com/} for supporting various open-source local LLM.
Inspired by the work of Reflexion~\cite{shinn2023reflexion}, we implement a simple reflective learning mechanism to learn from historical matches. To compensate for the limited numerical decision-making capacity of LLMs, we provide an identical simulator extracted from the game core to facilitate the future work such as search~\cite{yao2024tree} and counterfactual reasoning~\cite{fu2024preact}. We also offer a basic implementation of integrating the simulator into the agent framework.
To addressing the absence of in-game chatting in native gameplay, we develop a Discord-based chatbot interface  (Figure 1(f)) that connects directly with the game server. Players can engage in diplomatic negotiations with CivAgent via Discord\footnote{https://discord.com/}.
This project adheres to Unciv's MPL-2.0 open-source license and provides comprehensive documentation for further development as well as legal consent forms required from users for data collection purposes.

Overall, this paper contributes by developing CivSim and CivAgent, which provide researchers with a excellent testbed for establishing LLM-based human-like agents as digital players. We conduct preliminary benchmark experiments across multiple tasks. We successfully invite a hundred players to participate in a internal trial, validating the feasibility of building a low-cost data flywheel.

\section{Relate Work}
\label{sec2}

\subsection{Proxies of Human Behavior in Virtual Environments}
\label{sec2.1}
In the domains of human-computer interaction and artificial intelligence, a fundamental challenge is to create human-like agents that can emulate human behavior within interactive systems, akin to the portrayals seen in films such as ``The Terminator'' and ``Her''. 
Current investigations encompasses a variety of fields, including:
1) Believable agents~\cite{park2023generative, bates1994role} that emphasize consistency in language and behavior over extended periods;
2) Turing Test contenders~\cite{sejnowski2023large, zhang2022human,light2023avalonbench} like chatbots, which focus on achieving indistinguishability from humans through linguistic interaction;
3) Social simulations~\cite{hua2023war, pang2024self} that employ multiple agents to model the dynamics of social, economic, and political with the aim of conducting experiments not feasible in reality and gaining valuable insights.
In the gaming industry, agents in game were developed in the context of intelligent game non-player characters (NPCs)~\cite{riedl2012interactive,laird2001human} by rule-based approaches, such as finite-state machines~\cite{umarov2012believable, siu2021evaluation} and behavior trees~\cite{hecker2011my,pillosu2009coordinating}.
Reinforcement learning is also widely used in modern gaming to develop battle bots that far exceed human capabilities~\cite{vinyals2017starcraft, meta2022diplomacy}. 
Each of the aforementioned works reflects partial aspects of proxies of human behavior. In this paper, within an application, we can systematically study how to create human-like agents that can consistently behave in multi-turns human interactions, be difficult to discern as a machine, and make excellent decisions in challenging environments.


\subsection{LLM-based Autonomous Agents and Human Behavior}
\label{sec2.2}
With the substantial advancements in LLMs, there has been significant progress in the development of LLM-based agents for task planning and reasoning within interactive environments~\cite{zhang2023proagent,zhang2023building,zhu2023ghost,qi2023civrealm}. These agents have evolved to incorporate various architectures that complement the foundational LLM, such as retrieval-augmented generation~\cite{gao2023rag, gao2023retrieval}, memory strategies~\cite{park2023generative}, self-reflection mechanisms~\cite{yao2022react,shinn2023reflexion,madaan2024self}, learning capabilities~\cite{wang2023voyager,zhao2024expel,zhang2024large,shinn2023reflexion}, multi-agent workflows~\cite{hong2023metagpt, wu2023autogen}, and more. Given that LLMs encode a wide spectrum of human behaviors from their training data, some efforts have been made to develop human-like agents~\cite{park2023generative, light2023avalonbench,xu2023exploring}. Beyond the functional requirements of these agents, there is an emphasis on their capacity for human-computer interaction~\cite{bozkurt2023generative,wang2024survey}.
In this paper, inspired by the vision of ``digital employees'', we instantiate the concept of a ``digital player'' through the development of CivAgent. We draw upon current state-of-the-art architectures found in LLM-based agents but place emphasis on human-AI interaction and building data flywheels.

\subsection{Evaluation of Large Language Models and LLM-based Agents}
\label{sec2.3}
While all tasks can potentially serve as evaluation tools for LLMs, the question remains as to which can truly measure AGI capabilities~\cite{chang2024survey}. Existing LLM benchmarks can be categorized into knowledge and capability evaluation~\cite{hendrycks2020measuring, hendrycks2021measuring}, alignment evaluation~\cite{huang2023trustgpt,costa2023multilingual}, and robustness and risk Evaluation~\cite{zhu2023promptbench,wang2023plan}.  
In contrast to static evaluation protocols of LLMs, LLM-based agents are tested on open tasks within interactive environments. Among these, systematic benchmarks include API environments for tool learning~\cite{qin2023toolllm, schick2024toolformer, parisi2022talm}, text-based game environments~\cite{osborne2022survey,cote2019textworld,hausknecht2020interactive,urbanek2019learning} for language agent evaluation and multi-modal simulators for embodied agents~\cite{srivastava2022behavior,fan2022minedojo, reed2022generalist}.  CivRealm~\cite{qi2023civrealm}, a project built upon Freeciv (another similar but unpopular strategy game), primarily focuses on the comparison between LLMs and RL approaches in unit control and economic development.
The distinguishing feature of CivSim lies in its inherent requirement for human-AI interaction, allowing for a more systematic examination of an agent's capabilities from anthropomorphism to reasoning and planning. Most importantly, it offers the potential for low-cost construction of a data flywheel.

\section{Environment}
\label{sec3}
\subsection{Game Description}
\label{sec3.1}
In Unciv, players assume the role of civilization leaders with the objective to achieve victory through various paths such as science, diplomacy, and conquer. 
Over potentially hundreds of turns, players must engage in long-term planning across economic expansion, military development, and technological advancement while balancing short-term and long-term benefits. As the technology progresses, new buildings and units become available for production. The game encompasses 80 types of technologies, 68 types of buildings, 126 types of units, and six diplomatic states.
As civilizations develop unevenly and geopolitical landscapes shifts, diplomatic policies become crucial for ultimate victory. Players engage with each other through diplomatic actions such as declarations of war, peace negotiations, and alliances, coupled with employing natural language-based diplomatic dialogue (e.g., persuasion, commitments, and deception), to fulfill their objectives. As shown in Figure \ref{Fig:game}(a), Unciv offers a variety of convenient methods for the creation of new scenarios, such as generating random maps with diverse landscapes and varying numbers of players, or modifying difficulty levels of AI (solely impacting numerical bonuses).
In summary, Unciv presents unique challenges and complexities that make it an excellent testbed for LLM-based human-like agents. For further details about the game mechanics please refer to Appendix A.1.

\textbf{Observation.} We do not directly process the raw pixel data of the game interface, but extract representative discrete information from the game save file. These observations include data related to the map, units, cities, technologies, diplomacy, and overall scores. Map information includes whether specific tiles are visible, where ``tiles'' refer to the grid cells on a hexagonal map. The map provides detailed information such as terrain type, tile ownership, resource output, and more. Unit information offers insights into the position of units, their owners, health status, and whether they can take action. City information includes details about city location, ownership, population, and resource production. Technology information displays the technologies being researched and those already researched. Diplomacy information encompasses diplomatic relations with other participants. It includes status of friendliness, war, alliances, open border agreements, friendly declarations, research collaborations, ongoing trades, and a record of historical diplomatic events. The overall score consists of points in various aspects such as military and economic strengths. Dialogue history between civilizations is also part of the observations. In fact, dialogues will bring about more flexible diplomatic relations, such as future commitments and vassal relationships. Please refer to Appendix A.2.

\textbf{Actions.} We implement a comprehensive set of action interfaces covering four main aspects of the game: units, production, technology, and diplomacy. The unit action interface is responsible for controlling player units, including military and non-military units. Production actions involve the construction of buildings and units. Technology actions enable players to modify the technology they are currently researching. Diplomacy actions authorize players to propose a diplomatic agreement, such as scientific collaboration, declaring war, or forming alliances. Agents can initiate dialogue at any time to influence each other's decisions in these four aspects, and decisions change the game state through the action interfaces. Each diplomatic action defaults to initiating a diplomatic negotiation dialogue. We also provide switches for all action interfaces to allow the use of the native rule-based AI, which allows researchers to modify only a part of the AI system. For example, we can concentrate on diplomatic decisions and negotiations, disregarding production and military aspects. For a detailed list of implemented actions, please refer to Appendix A.3.
\subsection{Engineering Framework}
\label{sec3.2}
The native Unciv game is implemented by Kotlin and employs a rule-based artificial intelligence (AI) system. We architecturally separate the AI system from the core engine of game using a server-client mode, which allows for the AI module to be written in Python, a language more friendly to AI. For each AI-controlled civilization, when its turn arises, the game issues requests to the AI module via HTTP protocol. The civilization then acts based on the responses received. This design enables researchers to implement various algorithms for agent, including behavior trees and reinforcement learning. The only requirement is generating actions conforming to CivSim's specifications. since the native Unciv lacks a conversation function, we establish chatbots for each optional civilization within the game on Discord, which is a popular platform for both online gaming and AI applications. Upon initiating gameplay, players are required to transmit their game identification (gameid) to an administrative bot. A new Discord channel is then generated for group discussions and events broadcasting. It also allows private dialogues between civilizations. The gameid ensures synchronization between chatbots and accurate civilization within the certain match.

\textbf{LLM-friendly.} In CivSim, the limited number of agents significantly reduces the costs associated with invoking LLMs. As a turn-based game, CivSim does not require real-time responses. Such features aligns well with LLMs that typically necessitate considerable time for inference.

\textbf{Tools for data flywheels.}
To provide services for remote players and support the mode of multiple human players, a remote game host is needed. We develop a remote game server to replace the native server for interconnecting multiple game clients, Discord chatbots, and AI modules. Players can easily set the new URL of the remote server for access. We implement a data collection mechanism that allows players to rate the rationality of each AI decision. An administrative bot periodically invites players to score and evaluate the anthropomorphism of the agents. In future work, agents will masquerade as human players in aspects such as usernames to mitigate data bias. All related data are stored on the server. We develop tools for gathering save files and chat data every turn. Detailed documentation is provided to assist researchers in establishing their own game servers.

\subsection{Tasks}
\label{sec3.3}
Given a comprehensive action interface, agents can engage in the game as fully as human players. Due to the difficulty of achieving final victory conditions, we employ metrics provided by Unciv to quantify the advantage of civilizations at every turn. These include army strength, population strength, technology strength, and overall civilization strength—the latter being an aggregate indicator that synthesizes the others.
In addition to full game task, CivSim includes two mini-games that concentrate on specific facets of play: Diplomatic Negotiations and Deception. These mini-games leverage the competencies of CivAgent (i.e., maintaining same prompts) and can be executed independently from the game environment.
The Diplomatic Negotiations mini-game tasks players with engaging in diplomatic discourse aimed at securing advantageous positions. This task requires two agents to assume roles as buyer and seller respectively. They must infer their counterpart's bottom line while continually modifying trade proposals until a deal is struck. The metric for success is defined by where the final agreed price falls between both parties' real bottom lines (ranging from 0\% to 100\%).
In the Deception mini-game, participants are tasked with identifying deceptive content circulated by opponents regarding the game situation. The metric here is defined as the probability of opponents accurately identifying misinformation.


\begin{figure*}[tb]
  \centering
  \includegraphics[scale=0.15]{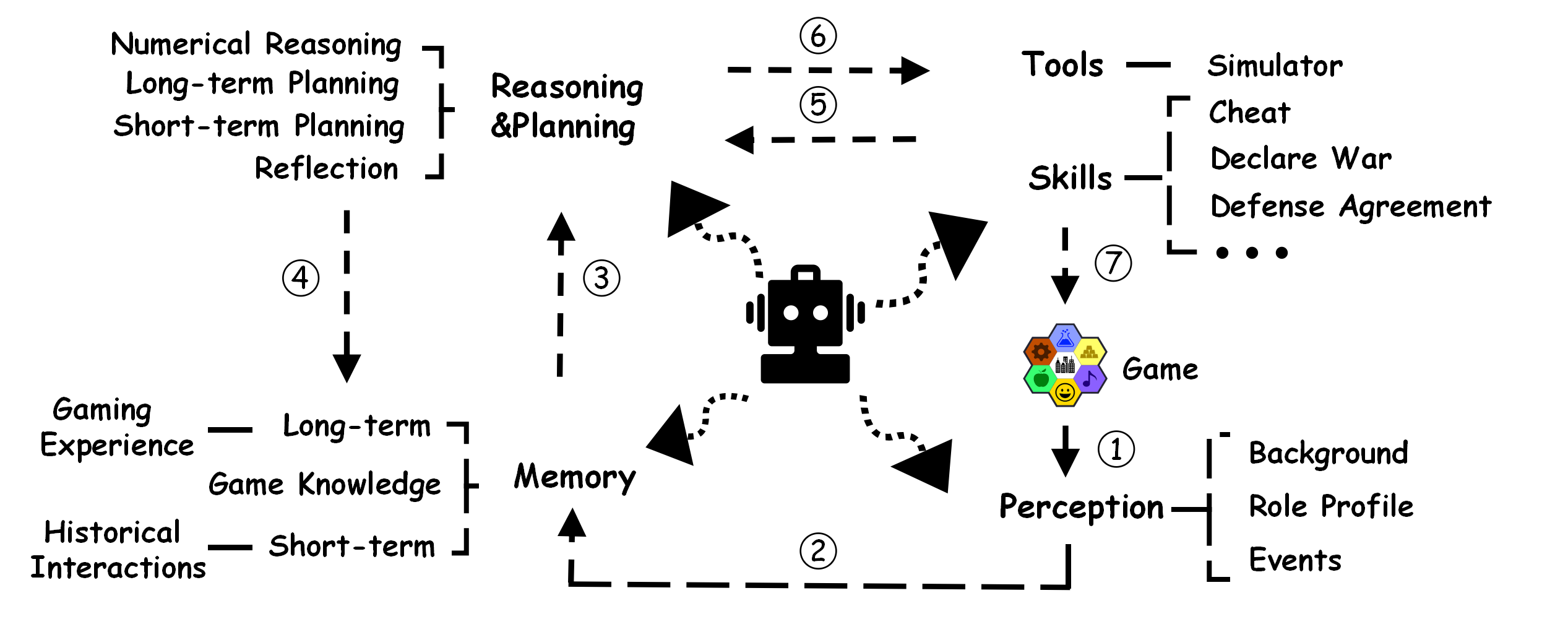}
  \caption{
  The architecture of the LLM-based CivAgent. It involves a decision-making process that can be delineated into the following steps:
\textbf{1.} Retrieve the current turn's game save file and extract observations;
\textbf{2.} Integrate game context from observations, dialogue history, historical events, etc., and employ RAG to retrieve experience required for prompts;
\textbf{3-6.} Based on observations and information provided by memory, CivAgent engages in reasoning, with permission to utilize tools like game simulators if necessary. Using this reasoned information, CivAgent formulates long-term and short-term plans within the skills spaces. After several turns or game end, CivAgent reflects on its  game situation and performance, and stores reflective experiences into long-term memory;
\textbf{7.} CivAgent then translates short-term planning (i.e., sequences of skills with filled parameters) into action interfaces for execution within the game environment.
  \label{Fig:Agent}
  }

\end{figure*}

\begin{figure*}[tb]

  \centering
  \includegraphics[scale=0.13]{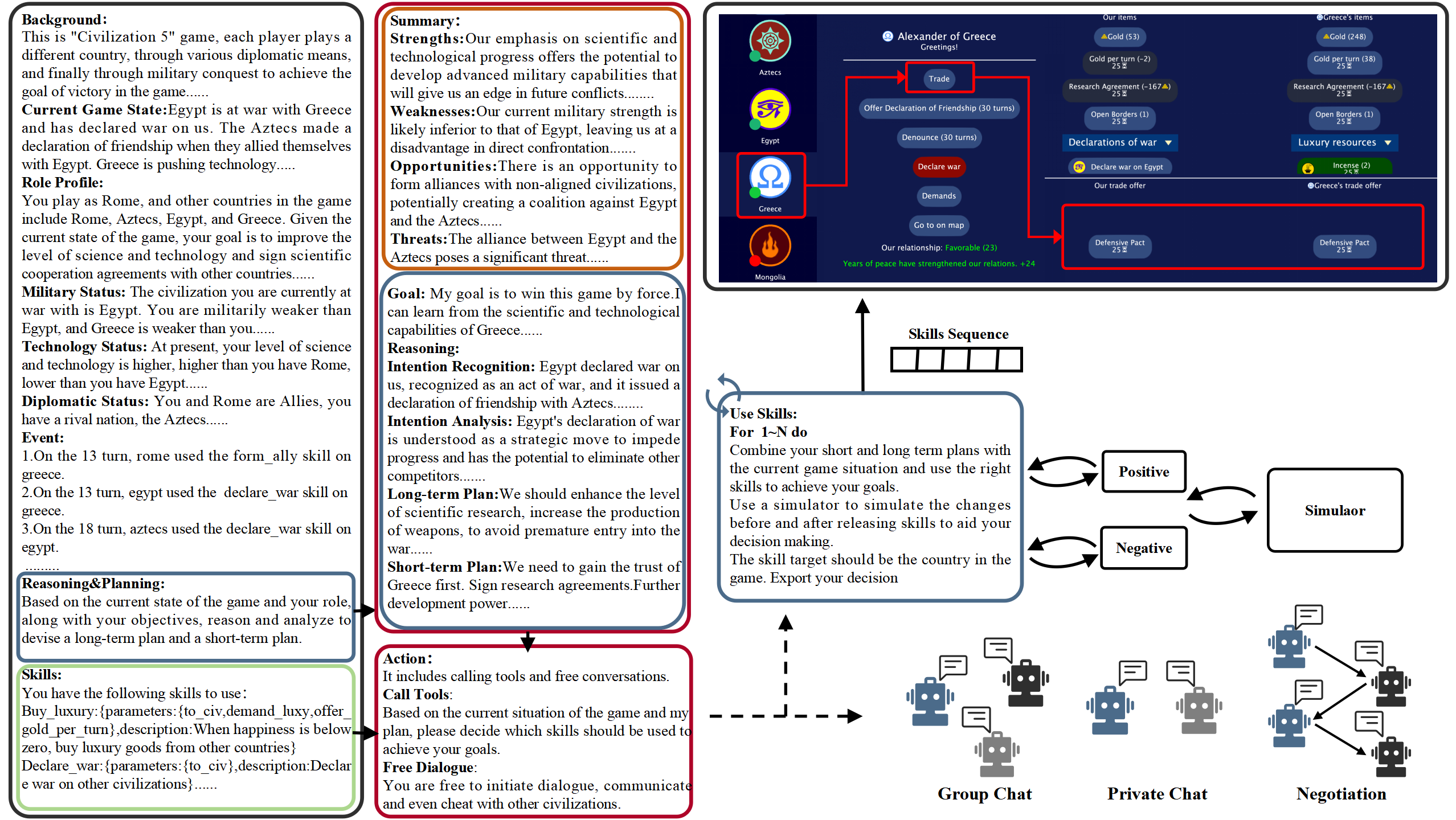}
  \caption{A simplified diagram of CivAgent from the perspective of the prompt.\label{Fig:prompt}}
\end{figure*}

\section{CivAgent}
\label{sec4}
To promote future research on digital players, we instantiate the concept of digital players as CivAgent within the CivSim environment. Following current state-of-the-art research of LLM-based agents, CivAgent is composed of several modules, as depicted in Figure \ref{Fig:Agent}.
To enhance the numerical reasoning capabilities of LLMs, CivAgent equips a game simulator which can simulate the outcomes of various decisions across any number of future turns. In the future work, the simulator could be replaced with a fully-trained sophisticated RL model. This approach also contributes to the generalizability of CivAgent since different games may simply require swapping out the corresponding simulators, thus reduce the need for game-specific prompt customization. It should be noted that while we provide action interfaces for units, our goal is not to achieve micro-management superiority like RL models might. Instead, CivAgent aims to function as a strategist or a diplomat who reasons and plans like humans do, gaining advantages at the strategic level. 
The following descriptions elaborate on each module:


 \begin{itemize}[leftmargin=0pt]
     \item\textbf{Perception Module.} As introduced in Section \ref{sec2.1}, the environment's observation consists of discrete information extracted from the game's save file. In unimodal mode, the observation is then organized into textual input through the prompt engineering. Figure \ref{Fig:prompt} depicted the prompt template used in CivAgent, including: 1) Background \textemdash providing a brief introduction to the Unciv game rules and the current game context; 2) Role Profile \textemdash informing the agent of the civilization it is currently playing and describing its own military, technological, and diplomatic status; 3) Events \textemdash the agent's recent 10 notifications (e.g., production or technology is completed) and the recent 20 worldwide events (e.g., news of wars or cities conquered).
     
     \item\textbf{Memory Module.} As shown in Figure \ref{Fig:Agent}, the memory module primarily stores data generated from the agent's interactions, including short-term memory, which consists of the most recent 20 lines of dialogue and actions, and long-term memory, which is generated by the reflection of skill usages and recent decisions. Basic memory strategies are also employed, including summarizing dialogue history beyond the range, and employing RAGs to recall experiences from the knowledge base.
     
     \item \textbf{Skills.} 
     In addition to free dialogue, we develop a range of skills for CivAgent that correspond to the action interface described in Section \ref{sec2.1}. Future work includes autonomously summarizing new skills like Voyager, involving code generation. As shown in Figure \ref{Fig:prompt}, we store the JSON format definitions of all skills within the prompt to enable LLM to select appropriate skills and generate corresponding parameters. These skills include 
1) Declare War: Initiate war against a civ;
2) Defense Agreement: Propose a defense pact with another civ.
3) Common Enemy: Solicit support from one civ to jointly attack another.
4) Seek Peace: Offer peace negotiations with a warring civ;
5) Research Agreement: Suggest signing a research agreement with another civ;
6) Change Closeness: Request an alteration in friendly relations with another civ;
7) Propose Trade: Initiate trade negotiations involving currency, cities, luxury goods, or various diplomatic treaties, involveing bargaining processes;
8) Production Priority: Provide production priorities between military and economic focuses;
9) Choose Technology: Choose a technology to research.

     \item \textbf{Reasoning and Planning.} As illustrated in Figure \ref{Fig:prompt}, the CivAgent engages in multi-step reasoning based on information provided by the perception and memory modules. It involves reasoning about opponents' actions and intentions, its own goals, and so on, which then informs both short-term and long-term planning. CivAgent employs a strategy akin to reflexion\cite{shinn2023reflexion} that allows for continuous learning from environments, and the experience are then stored in long-term memory. 
     
     \item \textbf{Tools.} To enhance LLM's numerical decision-making capabilities, we provide an game simulator as a tool for CivAgent, which is efficient and can simulate approximately 10 rounds within 1 second. It allows for the simulation of game while keeping diplomatic relations constant. As depicted in the top right corner of Figure \ref{Fig:prompt}, whenever the agent faces a decision, CivAgent can employ this simulator to forecast game outcomes across various decision choices after N rounds. The impact is represented as the changes in civilization strengths and other relevant metrics, and then incorporated into the prompt.
 \end{itemize}
 Due to space constraints, we defer the specific implementation of each module to Appendix B. Considering the delay and research interest, CivAgent focuses more on the diplomatic decision-making and negotiation aspects of Unciv, while retaining the native rule-based methods for unit control. CivAgent makes diplomatic decisions regarding whether to invoke skills every 5 turns, with a maximum of three skills. The recipients of these skills initiate negotiations on their own turns and ultimately respond with ``agree'' or ``disagree''. CivAgent brings approximately 5 seconds of extra inference time per turn compared to the native AI, which is deemed acceptable.

\section{Experiments}
\label{sec_experiment}
In this section, we demonstrate the potential of CivSim as a benchmark in the domain of LLM-based agents through a series of experiments. To facilitate reproducibility, we do not consider incorporating human evaluators into our assessment approach and solely utilize agents to participate in the full game task. All metrics used in our experiments are derived from quantitative data provided by environmental feedback. It is important to emphasize that CivSim's superiority is primarily reflected in its ability to attract real human participation and collect extensive human feedback data. For more details on the experiments, such as case studies, please refer to Appendix C.
\subsection{Baselines}
For the full game task described in Section \ref{sec3.3}, we employ GPT-3.5-turbo as the foundational LLM, and consider following variants of CivAgent to investigate the impact of different architectures:
\begin{itemize}[leftmargin=0pt]
    \item \textbf{Naive CivAgent(CivAgent-N):} The CivAgent variant that uses a single basic prompt.
    \item \textbf{CivAgent with basic workflows(CivAgent-W):} The most basic workflow, which breaks down decision-making tasks, does not involve simulation and reflection.
    \item \textbf{CivAgent with simulator (CivAgent-S):} The CivAgent variant that utilizes a workflow composed of multiple prompts, but does not include the simulator module and reflection module.
    \item \textbf{CivAgent with simulator and reflection(CivAgent-SR):} The CivAgent that includes all modules.
\end{itemize}
For mini-game tasks that primarily evaluate agents' skills in negotiation and deception, we employ the same agent workflows and consider the following Backbone LLMs:  GPT-3.5-turbo, GPT-4-1106-preview, Gemma-7B~\cite{team2024gemma}, Mistral-7B~\cite{jiang2023mistral} and  Llama3-8B~\cite{meta2024meta}.

\subsection{Experimental Settings}
In the full game setup, we initialize a total of 50 matches, each consisting of four fixed civilizations: Rome, Aztecs, Greece, and Egypt. Both the map and spawn locations are randomized for these encounters. For each match, we assign CivAgent-N, CivAgent-W, CivAgent-S, and CivAgent-SR to play one civilization each, exploring all 24 possible permutations of their arrangement. Matches are capped at 250 turns and ultimately scored based on civilization strength.
For the negotiation task, the dialogue is confined to a maximum of four rounds. In every experimental pair A-B for this task, LLM A assumes the role of buyer with a default trade proposal while LLM B plays seller and modify the initial trade offer.
Regarding the deception task: for each experimental pair A-B, LLM A acts as deceiver generating false information based on game context, meanwhile LLM B serves as detector assessing whether such misinformation was true or not. For the two mini-game tasks, CivAgent reuses the prompts used in full game tasks, and the results from human experts are also provided. Each pair performs ten repetitions to obtain an average result that would serve as a score for one trial session. These trials are then repeated across ten different save files twenty times.

\begin{table}[!ht]
\centering
\caption{Performance on the full game task under different agent architectures, where bold represents the best result, and $*$ denote the second best. 
``Freq.'' indicates how often agent uses a specific skill per game on average. ``SR'' refers to the success probability of diplomatic skills being accepted.}
\label{tab:1}
\resizebox{\textwidth}{!}{
\begin{tabular}{cccccccccccccc} 
    \toprule &
    & \multicolumn{2}{c}{Change Closeness} & \multicolumn{2}{c}{Seek Peace} & \multicolumn{2}{c}{Form Ally}&\multicolumn{2}{c}{Research Agreement}&\multicolumn{2}{c}{Common Enemy} &\multicolumn{2}{c}{Declare War}\\
    \cmidrule(lr){3-4} \cmidrule(lr){5-6} \cmidrule(lr){7-8}\cmidrule(lr){9-10}\cmidrule(lr){11-12}\cmidrule(lr){13-14}
    Methods & Avg. Score & Freq. & SR & Freq. & SR & Freq.& SR& Freq. & SR& Freq.& SR& \multicolumn{2}{c}{Freq.} \\
    \midrule
    CivAgent-N& 17.6 & 31.1 & 10.0\% & 25.6 & 48.0\% & 35.1 & 32.4\% & 11.9 & 72.7\% & 7.1 & 0.0\%& \multicolumn{2}{c}{19.1} \\
    CivAgent-W & 18.2 & 27.9 & 29.6\% & 22.2 & \textbf{54.6\%} & 30.2 & 44.8\% & 22.5 & 72.7\%  & 1.3 & 0.0\%& \multicolumn{2}{c}{9.8}\\
    CivAgent-S & 24.9$^{*}$& 23.9 & 30.4\%$^{*}$ & 15.3 & 50.0\%$^{*}$ & 29.9 & 48.3\%$^{*}$ & 16.1 & 73.3\%$^{*}$  & 9.8 & 11.1\%$^{*}$ & \multicolumn{2}{c}{12.9}\\
    CivAgent-SR& \textbf{39.2} & 22.1 & \textbf{38.1\%} & 11.5 & 54.6\% & 31.1 & \textbf{60.0\%} & 17.3 & \textbf{76.5\%} & 11.3 & \textbf{12.5\%}& \multicolumn{2}{c}{14.2}\\
    \bottomrule 
\end{tabular} 
}
\end{table}

\begin{table}[!ht]
\centering
\caption{Performance comparison between different LLMs on the negotiation task, with each row representing buyers (higher score is better) and each column representing sellers (smaller is better).}
\label{tab:2}
\scriptsize
\begin{tabular}{ccccccc|c}
\toprule
Model & GPT4 & GPT3.5 & Mistral & Gemma & Llama3 & Expert & Avg. Score \\
\midrule
GPT4 & 43.1($\pm$10.3) & 66.7($\pm$13.2) & 86.4($\pm$12.1) & 69.2($\pm$2.1) & 71.7($\pm$14.2) & 41.3($\pm$8.1) & 63.6 \\
GPT3.5 & 39.2($\pm$8.9) & 52.9($\pm$22.3) & 80.0($\pm$17.8) & 28.6($\pm$11.2) & 57.1($\pm$7.4) & 29.8($\pm$9.3) & 47.9 \\
Mistral & 34.8($\pm$12.6) & 41.2($\pm$3.9) & 59.1($\pm$16.8) & 35.7($\pm$6.1) & 54.7($\pm$22.1) & 28.8($\pm$7.2) & 42.4 \\
Gemma & 15.2($\pm$4.1) & 23.7($\pm$20.2) & 39.1($\pm$15.6) & 68.5($\pm$12.4) & 29.8($\pm$2.2) & 30.6($\pm$8.4) & 34.5 \\
Llama3 & 21.4($\pm$5.8) & 81.3($\pm$13.4) & 46.2($\pm$17.1) & 72.8($\pm$9.1) & 50.9($\pm$9.7) & 34.8($\pm$9.6) & 46.4 \\
Expert & 70.6($\pm$5.2) & 84.2($\pm$9.6) & 96.4($\pm$6.7) & 85.6($\pm$4.5) & 78.3($\pm$8.3) & N/A & \textbf{83.1} \\
\midrule
Avg. Score & 37.4 & 58.3 & 67.4 & 60.1 & 57.1 & \textbf{27.3} \\
\bottomrule
\end{tabular}
\end{table}

\begin{table}[!ht]
\centering
\caption{Performance comparison between different LLMs on the deception task, with each row representing deceivers (higher is better) and each column representing detector (smaller is better).}
\label{tab:3}
\scriptsize
{
\begin{tabular}{ccccccc|c}
\toprule
Model & GPT4 & GPT3.5 & Mistral & Gemma & Llama3 & Expert & Avg.\ Score \\
\midrule
GPT4 & 41.7($\pm$0.2) & 85.1($\pm$0.4) & 100($\pm$0.0) & 97.2($\pm$0.1) & 80.7($\pm$6.1) & 73.9($\pm$1.3) & 79.8 \\
GPT3.5 & 37.9($\pm$1.1) & 44.9($\pm$1.3) & 100($\pm$0.0) & 92.7($\pm$0.8) & 49.1($\pm$3.3) & 52.1($\pm$0.8) & 62.8 \\
Mistral & 28.5($\pm$0.2) & 70.1($\pm$2.5) & 100($\pm$0.0) & 97.1($\pm$0.4) & 86.5($\pm$1.1) & 44.8($\pm$2.1) & 71.2 \\
Gemma & 18.4($\pm$0.1) & 54.3($\pm$1.8) & 99.4($\pm$0.3) & 95.6($\pm$0.5) & 62.9($\pm$3.9) & 33.2($\pm$0.6) & 60.6 \\
Llama3 & 38.3($\pm$0.2) & 48.7($\pm$0.5) & 100($\pm$0.0) & 96.7($\pm$0.6) & 42.3($\pm$5.3) & 36.7($\pm$1.8) & 60.5 \\
Expert & 47.4($\pm$2.6) & 90.9($\pm$0.8) & 100($\pm$0.0) & 99.2($\pm$1.2) & 95.8($\pm$1.7) & N/A & \textbf{86.7} \\
\midrule
Avg.\ Score & \textbf{33.8} & 65.7 & 99.8 & 96.4 & 65.6 & 48.1 \\
\bottomrule
\end{tabular}
}
\end{table}

\subsection{Experimental Analysis}
Table \ref{tab:1} presents the average performance of different CivAgent variants in a four-player game setting. CivAgent-SR and CivAgent-S significantly outperform others, securing first and second place respectively, demonstrating the effectiveness of reflection and simulation modules in enhancing reasoning and planning capabilities. The table also shows the frequency and success rates of certain skills. Notably, CivAgent-SR achieves the highest number of successful ``Common Enemy'' and ``Form Ally'' actions, indicating its proficiency in diplomacy. It exhibits moderate frequencies for initiating ``Declare War'' and ``Research Agreement'', demonstrateing its good strategic balance between aggressive diplomacy and cautious engagement in warfare. A correlation is observed between high skill success rates and overall scores, implying that CivAgent-SR and CivAgent-S accurately assess game context to make insightful diplomatic proposals to other players. For the negotiation task, as shown in Table \ref{tab:2}, the most notable result is observed when GPT-4 functioned as the buyer and Mistral as the seller, achieving a buyer score of 86.4. Conversely, when Gemma served as the buyer and GPT-4 acted as the seller, it resulted in a minimal buyer score of 15.2. Averaging across rows provided an assessment of LLMs' capabilities as seller, while averaging across columns yielded evaluations for their performance as buyer. The data indicates that GPT-4 significantly outperforms other LLM baselines in both buying and selling roles, aligning with expectations. In negotiations against humans, however, LLMs still fail to fully master tactics such as maximum pressure and consequently underperform.
In the deception task presented in Table \ref{tab:3}, we apply the same analytical process and arrive at similar conclusions. An interesting observation is that GPT-4 scored higher than humans when acting as deceivers, which should be further validated .

\section{Limitations and Conclusion}
This paper aims to explore how to efficiently create interactive computational agents as human-like digital players within the context of a data-centric research paradigm. Inspired by citizen science games, we present CivSim, a unique testbed for LLM-based agents built upon the strategy game Unciv, which boasts millions of active players. 
This testbed places equal emphasis on both agent capabilities and anthropomorphism. 

While this study offers significant insights into the development of agents, it is crucial to recognize its limitations, including the restriction to a single gaming scenario and the absence of experiments on how human feedback data can be further utilized to enhance agent performance. Future research should explore various efficient methods for data utilization. 
The practices on this project will facilitate further applications of LLM-based agents in various domains.

\newpage
\bibliographystyle{abbrvnat}
\bibliography{main}

\newpage
\appendix

\section*{Checklist}
\begin{enumerate}

\item For all authors...
\begin{enumerate}
  \item Do the main claims made in the abstract and introduction accurately reflect the paper's contributions and scope?
    \answerYes{See Section~\ref{sec2}.}
  \item Did you describe the limitations of your work?
    \answerYes{}
  \item Did you discuss any potential negative societal impacts of your work?
    \answerNo{Our method does not have negative social impacts.}
  \item Have you read the ethics review guidelines and ensured that your paper conforms to them?
    \answerYes{}
\end{enumerate}

\item If you are including theoretical results...
\begin{enumerate}
  \item Did you state the full set of assumptions of all theoretical results?
    \answerNA{}
	\item Did you include complete proofs of all theoretical results?
    \answerNA{}
\end{enumerate}

\item If you ran experiments (e.g. for benchmarks)...
\begin{enumerate}
  \item Did you include the code, data, and instructions needed to reproduce the main experimental results (either in the supplemental material or as a URL)?
    \answerYes{See \textbf{ \textit {\url{https:/github.com/fuxiAIlab/CivAgent}}}.}
  \item Did you specify all the training details (e.g., data splits, hyperparameters, how they were chosen)?
    \answerYes{See Section~\ref{sec_experiment}.}
	\item Did you report error bars (e.g., with respect to the random seed after running experiments multiple times)?
    \answerYes{See Section~\ref{sec_experiment}.}
	\item Did you include the total amount of compute and the type of resources used (e.g., type of GPUs, internal cluster, or cloud provider)?
    \answerYes{See Appendix C.}
\end{enumerate}

\item If you are using existing assets (e.g., code, data, models) or curating/releasing new assets...
\begin{enumerate}
  \item If your work uses existing assets, did you cite the creators?
    \answerYes{We cited.}
  \item Did you mention the license of the assets?
    \answerYes{See Section~\ref{sec1}.}
  \item Did you include any new assets either in the supplemental material or as a URL?
    \answerYes{See \textbf{ \textit {\url{https:/github.com/fuxiAIlab/CivAgent}}}.}
  \item Did you discuss whether and how consent was obtained from people whose data you're using/curating?
    \answerYes{See Section~\ref{sec1}.}
  \item Did you discuss whether the data you are using/curating contains personally identifiable information or offensive content?
    \answerNA{No identifiable information.}
\end{enumerate}

\item If you used crowdsourcing or conducted research with human subjects...
\begin{enumerate}
  \item Did you include the full text of instructions given to participants and screenshots, if applicable?
    \answerNA{}
  \item Did you describe any potential participant risks, with links to Institutional Review Board (IRB) approvals, if applicable?
    \answerNA{}
  \item Did you include the estimated hourly wage paid to participants and the total amount spent on participant compensation?
    \answerNA{}
\end{enumerate}

\end{enumerate}
\clearpage
\section{Environment}
\begin{figure*}[tb]
  \centering
  \includegraphics[scale=0.2]{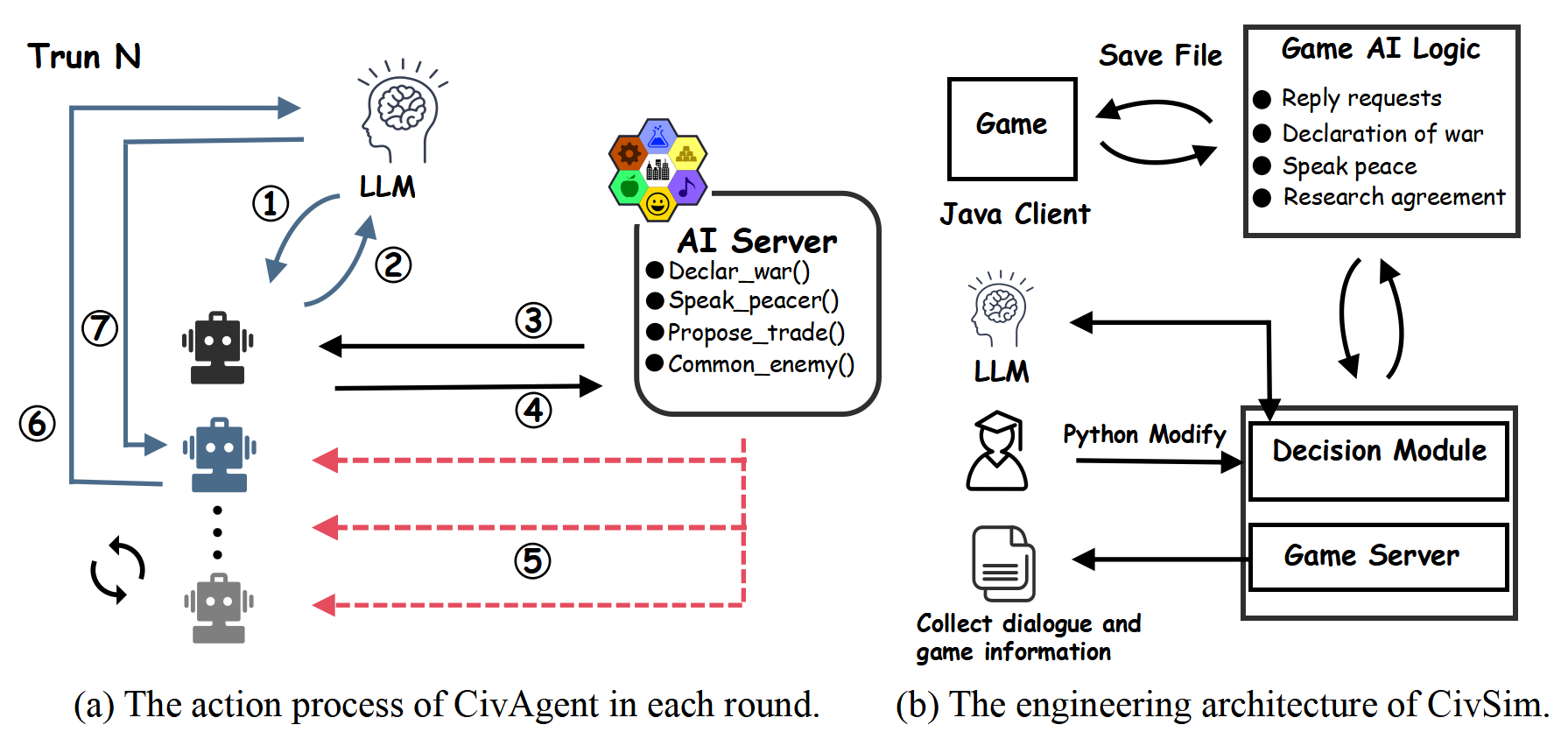}
  \caption{(a) A schematic diagram illustrating the procedural steps for CivAgent's actions in each round, where robots of different colors represent different civilizations. The steps are \textbf{1-2.} Using LLM to determine which skills to activate; \textbf{3-4.} The game server calls the agent to use skills; \textbf{5.} If there is a decision request, it is sent to the corresponding agent; \textbf{6-7.} Use LLM to make decisions about requests. (b) The engineering architecture diagram of CivSim. The game client synchronizes data with the AI module through a save file. The AI module makes decisions on different aspects in a specific order. The CivAgent and multiplayer game server, coded in Python and easily modifiable, are shown in the bottom right corner of the diagram.\label{Fig:action}}
\end{figure*}
In this section, we provide a more detailed description of the CivSim environment, including additional information on the quantification of evaluation metrics and minor modifications to the native Unciv game rules that we have implemented. Figure ~\ref{Fig:action} serves as an important supplement to Section 3.2, which discusses the engineering architecture of CivSim.

\subsection{Detailed Descrption of Full Game }
Unciv is a strategy game that simulates the development process of civilization in the real world. Prior to starting gameplay,each player chooses to play as a civilization leader with each civilization possessing distinct capabilities known as ``Unique Abilities(UA)'', along with special military units and building. At the first round, the player needs to maneuver settler units to choose a optimal location on the map for establishing their capital city, considering resource distribution, terrain, and proximity to other civilizations. Once a city has been founded, it can construct various buildings and units. Buildings such as farms, mines, libraries, and opera houses serve to augment population growth, productivity levels, scientific research capacity, and cultural development, respectively. Notably among these are wonders --- worldwide unique buildings with powerful effects.
In early turns,player may direct scout units to explore uncharted territories on the map in order to discover other players' capital, new resources and natural wonders. Subsequently, the newly produced settler units may establish new cities in favorable locations previously scouted thereby enhancing civilization strength.
If neighbors are pacifists, player can further develop civilization strength through 1) constructing worker units to improve tiles thus improving food output and production output of cities; 2) investing research points to upcoming technologies that unlock new units and buildings; 3) cultivating culture and religion, the former potentially unlocking governmental policies yielding various bonuses while latter could entail founding spreading religious beliefs garnering doctrinal benefits.
Conversely if neighbors exhibit belligerence,defensive measures become imperative,inlcuding recruiting troops, building fortifications and researching military technology.
Every building and unit will consume gold as maintenance cost per turn, and gold can also be used as a general equivalent to directly purchase units and buildings. Therefore, economic development is also crucial.
As the game progresses, diplomacy gains increasing importance. The supported treaties within game includes alliances (mutual defense pacts), open border agreements, friendly declarations, research collaborations, ongoing trades.
This is the basic complete game flow of Unciv,Its complexity is reflected in the mutual influence of various numerical systems and the inherent game-theoretic nature of competition among multiple civilizations,presenting significant challenges to current LLM-based agents. Below, we briefly describe the various systems of the game as follows:

\begin{itemize}[leftmargin=0pt]
\item \textbf{Victory:} There are multiple paths for victory: 1) Culture: complete the 5 social policy branches and utopian projects; 2) Domination: conquer all other civilizations and become the last surviving civilization; 3) Technology: complete the construction of a spaceship to Centaur $\alpha$; 4) Diplomacy: establish the United Nations and win the final election. These different conditions put forward high requirements for agent's strategy preference and strategic planning ability.
\item \textbf{City and Construction:} 
A city is capable of constructing an array of buildings and units. Each city possesses a population that is allocated across tiles to generate food and production point, where food accumulates to support additional population growth, and production is utilized in the construction of various buildings and units. Worker units can be employed to imporove tiles --- for instance, building farms on plains to enhance food yield of tiles. Tiles also provide strategic resources necessary for certain buildings and military units, as well as luxury resources which constrain population growth within a civilization. The advancement of certain technologies, such as Fertilizer, can further increase tile output. Buildings serve as pivotal nodes connecting the production system with other systems like science and culture. Most science and culture points are generated by buildings.

\item \textbf{Science and Technology:} As the upon the establishment of the city, a technology system can be unlocked. This system is segmented into nine eras ranging from the ancient era to the future era and is represented by a tech tree composed of 80 major discoveries and innovations in human history. Players utilize research points produced by their cities to study technologies, thereby unlocking new buildings, units, and subsequent technologies,connecting to other systems and causing impacts. Just like in the real world, research plays the most crucial role in the development of civilizations in the game.
\item \textbf{Units and Warfare:} Units are categorized into military and non-military types, with the former primarily engaged in combat operations and the latter consisting of workers, great persons, and other productive units. Military units possess attributes such as attack strength, defense strength, and health. Each unit is allotted one move and one attack per turn. Combat damage is calculated automatically by the game engine. A unit vanishes when its health falls below zero. The concept of conquest in the game is predominantly manifested through the capture of cities --- when a city's defense points drop to zero, both the city itself and its surrounding tiles fall under control of the attacking civilization. A common strategy of Unciv is increasing a civilization’s number of cities and overall output through conquest rather than development.

\item \textbf{Culture and Religion:} In addition to the science system, the game also possesses culture and religion systems. Similar to science, these systems rely on corresponding outputs from city buildings and unlock specific game bonus through the accumulation of culture points and faith points. For instance, the ``republic" policy can provide an additional production point for each city. Through such game bonus, the development of culture and religion can further enhance scientific progress, city productivity, and even military strength.

\item \textbf{Diplomacy:} 
In native Unciv, diplomatic relations between civilizations are primarily determined by various events that affect their friendly relations, such as prolonged periods of peace, which can increase the value of friendly relations. Civilizations proactively initiate diplomatic events based on certain rules, such as refraining from attacking distant civilizations. Cooperation and competition between civilizations have a significant impact on the game situation. Cooperation, such as defensive pacts, allows for collaborative opposition against military powers, while research collaboration agreements can significantly accelerate technological development (notably, players with a lack of research collaboration agreements may lag behind internationally). In terms of competition, military powers may threaten weaker civilizations by demanding gold and may also invite third parties to jointly attack other civilizations. The diplomatic interface also provides a powerful trading system supporting any combination of any number of elements, including gold, resources, cities, diplomatic treaties, and more.

\end{itemize}

Regarding a more detailed introduction of each system of Unciv, you can refer to \textit{\textbf{https://civilization.fandom.com/wiki/Unciv}}. During the development of CivSim, we made some modifications to the native Unciv to better adapt it for research purposes, please see Appendix A.6.

\subsection{Observation}
Rather than directly utilizing the raw pixel information of game interfaces, CivSim extract discrete and representative information from game save files to achieve a deep understanding of game states as the environmental observation.
The save file of Unciv can reconstruct an entire game across clients. Thus, it contains all pertinent contextual information. Table \ref{tab:4} details the save file, including map, unit, city, technology, diplomatic relations, and civilization scores. The map and unit information, including each unit's health points, attack strength, defense strength, and coordinates, as well as the terrain and detailed information of nearby coordinates, can indicate which civilization has the advantage on the battlefield and enable the agent to perform tactical maneuvers for each unit.
The city information records population, productivity, existing buildings, and more for each city, enabling the agent to make better decisions regarding economic development and what each city constructs. Diplomacy and civilization score information document existing diplomatic treaties and score comparisons between civilizations, allowing the agent to understand the international situation and devise favorable diplomatic strategies. As the number of civilizations, cities, and units is not fixed, information such as civilizations, cities, and units are all variable-length dictionaries.
When performing information extraction, it's important to note that this is an incomplete information game. Each civilization should only obtain detailed information about its own civilization, the civilization scores of other countries, and the unit information of other countries within the line of sight.
Overall, these environmental information provides agent with a comprehensive perspective on the gaming context thereby facilitating numerical reasoning and long-term planning.

\begin{table}[!t]
\centering
\caption{Observations from the save file.}
\label{tab:4}
\scriptsize
\resizebox{\textwidth}{!}
{ 
        \begin{tabular}{cccc} 
            \toprule 
            Fields & Attributes & Value domains & Descriptions  \\
            \midrule
            & X & [0,M]& X-coordinate\\
            & Y & [0,N]& Y-coordinate\\
             & Base Terrain& [0,8]&Original topography\\
             & Terrain Features &[0,22] & Characteristic of terrain\\
              Map & Explored By& text & Civilizations that have explored this tile\\
             & Resource &[0,34] &Resources owned by this post\\
             & Road Owner & text & The owner of the map\\
             & History &  text &Historical information about the tile\\
            \midrule
            & Owner &text & The current owner of the unit\\
             & Original Owner& text&The original owner of the unit\\
             Unit & Name & text& Name of the unit\\
             &Health& [0,65535]&Unit blood volume\\
            & Promotions&[0,103] & Unit upgrades\\
             & Movement Memories & [0,2\textsuperscript{16}-1] &Unit historical movement\\
            \midrule
            & X & [0,M]& X-coordinate\\
            & Y & [0,N]& Y-coordinate\\
             & Id& [0,16\textsuperscript{32}-1] &City ID\\
             &Health& [0,2\textsuperscript{16}-1]&City blood volume\\
             City & Name &text & Name of the city\\
            & Founding Civ& text & Civilization to which it belongs\\
             & Population & [0,2\textsuperscript{16}-1] &Population and food quantity\\
             & City Constructions& [0,2\textsuperscript{16}-1] &Urban Construction Information\\
             &Expansion & [0,2\textsuperscript{16}-1] &Civilization storage number\\
             &Religion& text &Information on urban religion\\
             &Espionage & [0,2\textsuperscript{16}-1] &Espionage in the city\\
             &Tiles & Dict &All tile locations of the city and locations where work is being done\\
             &Is Original Capital& 0 or 1&Capital or not\\
             &Demanded Resource& [0,34]&Resources the city needs\\
             &Connected To Capital Status& 0 or 1&Whether it is connected to the capital\\
            \midrule
            & Science Of Last 8 Turns & [0,79]& Information on the last eight rounds of research\\
             & Techs Researched& [0,79] &Technology that has been studied\\
             Technology & Techs To Research & [0,79]& Current research projects\\
            & Techs In Progress& [0,2\textsuperscript{16}-1]& Current Research Progress\\
            \midrule
            & Other CivName & text & The name of diplomatic civilization\\
             & Trades& Dict &Trade information (including open borders, research agreements,etc.)\\
             Diplomacy & Peace & 0 or 1& State of diplomacy with this civilization\\
             & Protector & 0 or 1& State of diplomacy with this civilization\\
             & War & 0 or 1& State of diplomacy with this civilization\\
             & DefensivePact & 0 or 1 & State of diplomacy with this civilization\\
            & Flags Countdown& [0,2\textsuperscript{16}-1] & The number of rounds remaining in a diplomatic treaty with that civilization\\
            & Diplomatic Modifiers& text & Quantifies the diplomacy score with that civilization\\
            \midrule
             & Gold&[0,2\textsuperscript{16}-1] & All the gold of civilization\\
             Civilizations& Civ name&text &The Name of Civilization\\
             &Proximity & text &Distance from other civilizations\\
             &Natural Wonders& [0,2\textsuperscript{16}-1] &The civilization has natural attractions\\
             &Resource&  [0,2\textsuperscript{16}-1] &The type and number of resources a civilization has\\
             &Notifications & text &The history of civilization informs information \\
            \midrule
            & S & [0,2\textsuperscript{16}-1] &Civilization total score\\
            & N & [0,2\textsuperscript{16}-1] &The total population of a civilization\\
            & C & [0,2\textsuperscript{16}-1] &Crop production of civilization\\
            & P & [0,2\textsuperscript{16}-1] &Civilized production capacity\\
            Score& G & [0,2\textsuperscript{16}-1] &The total number of gold in civilization\\
            & T & [0,2\textsuperscript{16}-1] &The size of a civilization's territory\\
            & F &[0,2\textsuperscript{16}-1]  &Civilized military power\\
            & H & [0,2\textsuperscript{16}-1] &The happiness of civilized inhabitants\\
            & W & [0,2\textsuperscript{16}-1] &The level of science and technology of civilization\\
            & A &[0,2\textsuperscript{16}-1]  &A civilized level of culture\\
            \bottomrule
        \end{tabular}
    }
\end{table}

\subsection{Action}
We have developed a comprehensive set of action interfaces, as shown in Table \ref{tab:5}, which encompass four key aspects of gameplay: unit control, production management, technology development, and diplomacy. As depicted in Figure \ref{Fig:action}, the game sends HTTP requests to the AI module through these interfaces, and the game engine then processes the responses from the AI module to effectuate corresponding changes within the game.
The Unit Action interface enables precise manipulation of player units by setting destination coordinates for their movement. Additional action interfaces that allow for detailed control over each unit's combat behavior are currently under development. The Production-related interface grants agents complete authority over construction decisions in each city. This empowers agents to formulate economic development plans intelligently based on current contextual factors and strategic requirements within the game.
Interfaces related to Technology Actions permit agents to determine which technological advancements are most pertinent to their civilization"s progress and prioritize research accordingly. 
Diplomacy-related action interfaces are also invoked every turn to decide whether diplomatic proposals should be initiated.

\begin{table}[!ht]
\centering
\caption{The action interfaces supported by CivSim.}
\label{tab:5}
\scriptsize
\resizebox{\textwidth}{!}
{ 
        \begin{tabular}{cccccc} 
            \toprule 
            Type & Name & Descriptions & Input & Output & Related Skills \\
            \midrule
            Unit&tryHeadTowardsEnemyCity &set the target coordinates for unit&unit\_id:Int & Target\_coordinates:String & N/A\\
            \midrule
            Production & chooseNextConstruction & Select the next building to construct & target\_civ:String& Construction:String&Production\_Priority\\
            \midrule
            Technology & chooseTechToResarch & Choose the technology to study & target\_civ:String& Tech:String& Choose\_Technology\\
            \midrule
            Diplomacy & wantsToSignDeclarationOfFrienship&Whether to propose a friend declaration& target\_civ:String&Result:Bool &Change\_Closeness\\
           Diplomacy&offerPeaceTreaty&Whether to propose a peace treaty& target\_civ:String&Result:Bool & Speek\_Peace\\
           Diplomacy&declareWar& Whether to declare war& target\_civ:String&Result:Bool & Declar\_War\\
            Diplomacy&wantsToSignDefensivePact& Whether to sign a mutual defense pact & target\_civ:String&Result:Bool & Defense\_Agreement:\\
            Diplomacy&canSignResearchAgreementsWith&Whether to sign a research collaboration& target\_civ:String&Result:Bool & Research\_Agreement\\
              Diplomacy&getTradeAcceptability&Whether to approve the trade request& target\_civ:String& Result:Bool & Decision\_Trades\\
            Diplomacy&potentialLuxuryTrades&Offer to buy luxury goods& target\_civ:String& Trades:Dict & Propose\_Trade:\\
            Diplomacy&proposeCommonEnemy&An invitation to attack together& target\_civ:String& Trades:Dict & Common\_Enemy\\
            Diplomacy&proposeTrades&Making a transaction request& target\_civ:String& Trades:Dict & Propose\_Trades\\
            Diplomacy&cheat&Spread misinformation to others& target\_civ:String& Message:String & Cheat\\

            \bottomrule
        \end{tabular}
    }
\end{table}

\subsection{Evaluation Metrics}
CivSim adopts the civilization assessment metrics from native Unciv, which provides scores across 10 dimensions to evaluate civilization performance, including population, food, production, gold, territory area, military strength, happiness, technology, culture, and an overall performance assessment known as civilization score.
To ensure score balance across maps of different sizes, Unciv first calculate a map size adjustment factor, denoted as $m$. Taking the tile count of a medium-sized map as the baseline at 1276 tiles, if the current map's tile count $t$ exceeds this baseline, $m$ will be greater than 1, thus providing a positive bonus to the score. Let the number of cities be denoted as $c$, total population as $p$, wonder count as $w$, technology count as $s$, and future technology count as $f$, and military strength as $k$, the civilization score is defined as:

\begin{equation}
\text{Score} = (c \times 10 \times m) + (p \times 3 \times m) + (t \times 1 \times m) + (w \times 40 \times m) + (s \times 4) + (f \times 10) + (k \times 0.1)
\end{equation}

And the military strength $k$ is defined by the following formula:
\begin{equation}
k =  \left( \sum_{u \in \text{units}} (u \cdot {\text{isWaterUnit}(u)} \text{?} \frac{1}{2}:1 )\right) \times \min \left(\text{goldBonus}(g), 2 \right)
\end{equation}

Here, $u$ represents a unit within the civilization, and $units$ is the set of all units owned by the civilization. $isWaterUnit(u)$ returns 1 (true) if the unit $u$ is a water unit, and 0 (false) otherwise. $goldBonus(g)$ is a bonus factor calculated based on the civilization's gold $g$. It is derived by taking the square root of the gold amount, converting it to a percentage, and capped at a maximum value of 2.

This formula takes into account the impact of city development, population size, tiles, wonders, technology, and future technology development on civilization scores. The introduction of the map size adjustment factor $m$ ensures fairness in score calculation across different map sizes, thus making the game scoring system more precise and comprehensive.

\subsection{Detailed Descrption of Mini Game}
In addition to the full game tasks, CivSim also includes two mini-games that focus on specific aspects of the game: diplomatic negotiations and deception. Below are the processes of the two mini-games.

\textbf{Diplomatic Negotiations Minigame.} In the initial phase of the game, buyers initiate trade negotiations with a preset transaction proposal, such as attempting to exchange 20 gold for 2 luxury gemstones. Buyers must determine their maximum acceptable price based on the proposed transaction content --- that is, the highest amount they can pay without incurring a loss to their civilization score. To this end, we have devised a binary search strategy based on the simulator. The search interval is confined between the initial trading price and a predetermined maximum value. By utilizing simulators to emulate changes in civilization score after 20 rounds, we are able to identify a point at which the benefits gained from acquiring 2 luxury gemstones offset the loss of golds.
Sellers similarly employ this method to ascertain their minimum acceptable price. Once both parties have determined their bottom lines, we inform each party of market reference prices fluctuating within $\pm$ 20\% of these true bottom lines --- this serves as an approximation under conditions of imperfect information during negotiations.
Then the seller responds to the initial proposal based on the workflow of CivAgent. In each round, it is assessed whether or not the trade has been successful; if deemed unsuccessful, negotiations continue. If no agreement is reached after 4 rounds, then it is concluded that trade has failed.
If a successful transaction occurs, we can measure the negotiating ability of the agent by calculating how closely the final trading prices align with either party's bottom line. The metric here is defined by where the final agreed price falls between both parties' real bottom lines (ranging from 0\% to 100\%).

\textbf{Deception Minigame.} First, the deceiver creates and disseminates false information based on the gaming context. Upon receiving this information, the detector who shares the same information with deceiver initiates a process to determine its veracity. Below are examples of prompt instructions: For Deceiver, the heuristic task description used in the prompt is ``You now need to fool [\textit{civname}] with the current situation in order to win the game.''.
For Detector, it is ``[\textit{civname}] says [\textit{text}] to you. They may be cheating on you, or they may be sincere, and you need to combine the game situation to determine their true intentions.''.
An example of generated misinformation is ``Dear Egypt, I believe that our current peaceful relations provide a great opportunity for both of our countries to prosper. I suggest that we make a scientific research agreement to further enhance our technological capabilities and strengthen our military power. Additionally, I propose that we engage in a cultural exchange to share our rich heritage and foster better understanding between our peoples. Together, we can surpass all potential threats and emerge as the dominant forces in this game. What do you think?''
If identified as true, then for that round, victory is awarded to the deceiver; if recognized as false, then success belongs to the detector. The proportions of each outcome will be tallied and the successful rate of deceiver is used as an evaluative metric.

\subsection{Additional Modifications on CivSim}
We have made the following modifications to the CivSim to better serve as a benchmark test for evaluating agent capabilities:
\begin{itemize}[leftmargin=0pt]
\item We have altered the victory condition to be achievable solely through conquest by warfare.
\item Diplomatic actions that originally required technological unlocks, such as research agreements and mutual defense pacts, have been made available from the beginning.
\item In order to assess the flexibility and consistency of AI diplomatic policies, we have reduced the minimum number of turns for civilizations to make peace after declaring war from 10 turns to 1 turn, and the minimum interval between declaration of war from 10 turns to 1 turn.
\item We have introduced new rules to mitigate the advantage of conquest strategies in small maps: 1) For every military unit produced, the population of the city is reduced by one; 2) Healing for soldiers stationed in cities has been adjusted to +10, within borders +3, and no healing outside the borders; 3) The healing bonus for unit upgrades has been changed from +50 to +20; 4) Workers can now modify a tile up to 3 times, instead of an unlimited number of times.
\end{itemize}

\section{More Descrption of CivAgent}
\subsection{Tools Module}
\begin{figure*}[t]
  \centering
  \includegraphics[scale=0.18]{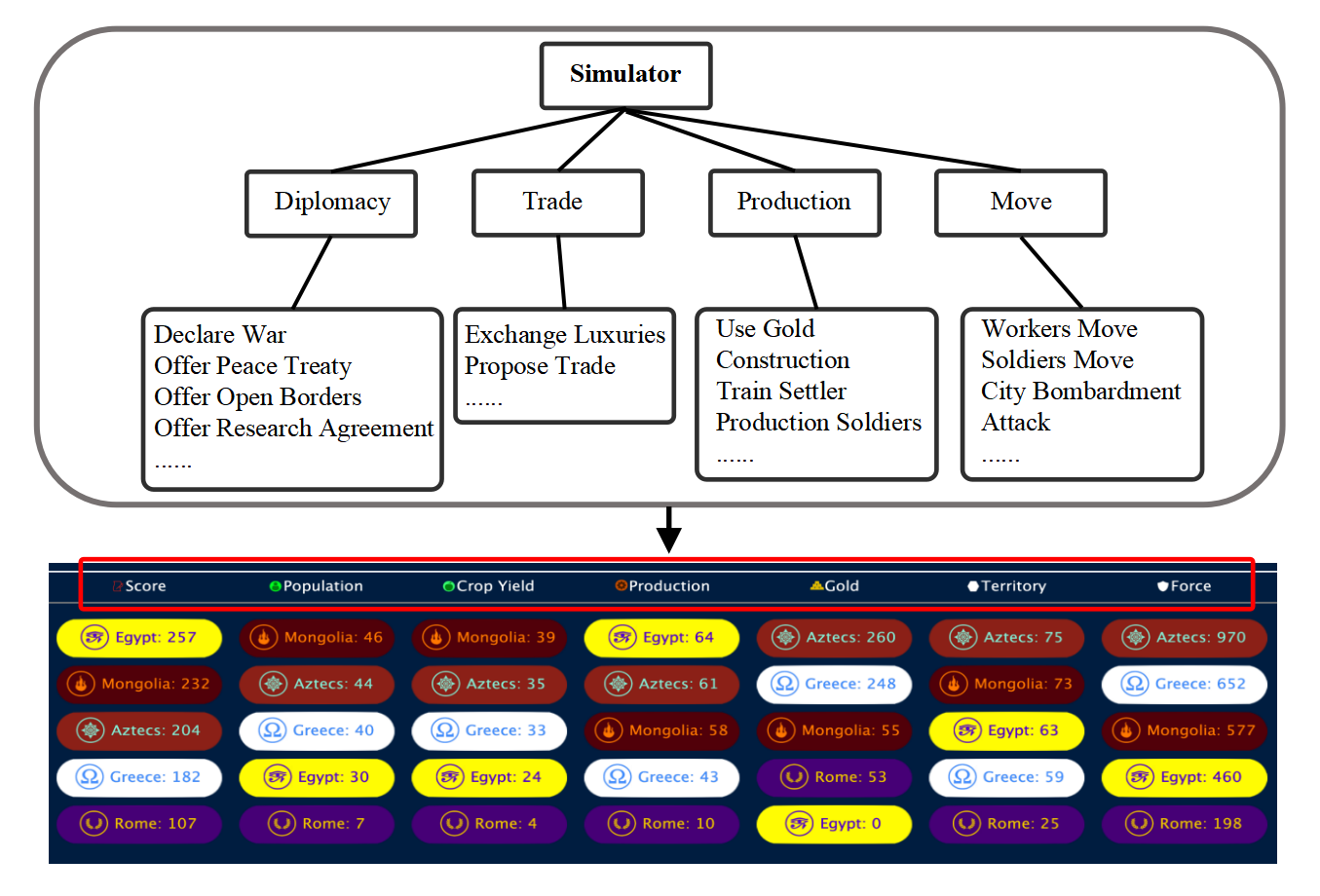}
  \caption{ Diagram of decision logic involved in the simulator. \label{Fig:simulation}}
\end{figure*}

In the tools module, inspired by AlphaZero's rollouts strategy, we developed a game simulator to rapidly evaluate the agent's decisions. This simulator is driven by the rule-based AI strategy of native Unciv, taking a save file as input and providing the save file after simulating N turns as output. The simulator's parameters include 1) ``turns'', indicating how many turns to run; 2) ``diplomacy switch'', to maintain consistent diplomatic relationships during the simulation, preventing additional events like  wars; 3) ``worker switch'', which can be toggled off in the game to speed up simulations, as worker-related simulations tend to be time-consuming in the rule-based strategy. 
As depicted in Figure \ref{Fig:simulation}, the simulator can simulate various aspects of the game, including diplomatic, trade, production, and warfare scenarios. For instance, each time the agent attempts to make a declaration of war, the simulator can be used to simulate the game state both before and after the declaration. The resulting game states after N turns can be employed to assess whether the decision has a positive or negative impact, as well as to what degree, and can be further integrated into the prompt.

In this manner, the simulator serves not only as an evaluative tool to measure the potential impact of decisions but also provides a valuable reference framework, aiding in understanding the long-term consequences of decisions made by the large model.  This enables decision-makers to adjust and optimize strategies based on more comprehensive and in-depth analyses, ensuring that each decision lays a solid foundation for the ultimate victory in the game.  

\subsection{Workflows of CivAgent}
\label{workflow}
\begin{figure*}[tb]
  \centering
  \includegraphics[scale=0.16]{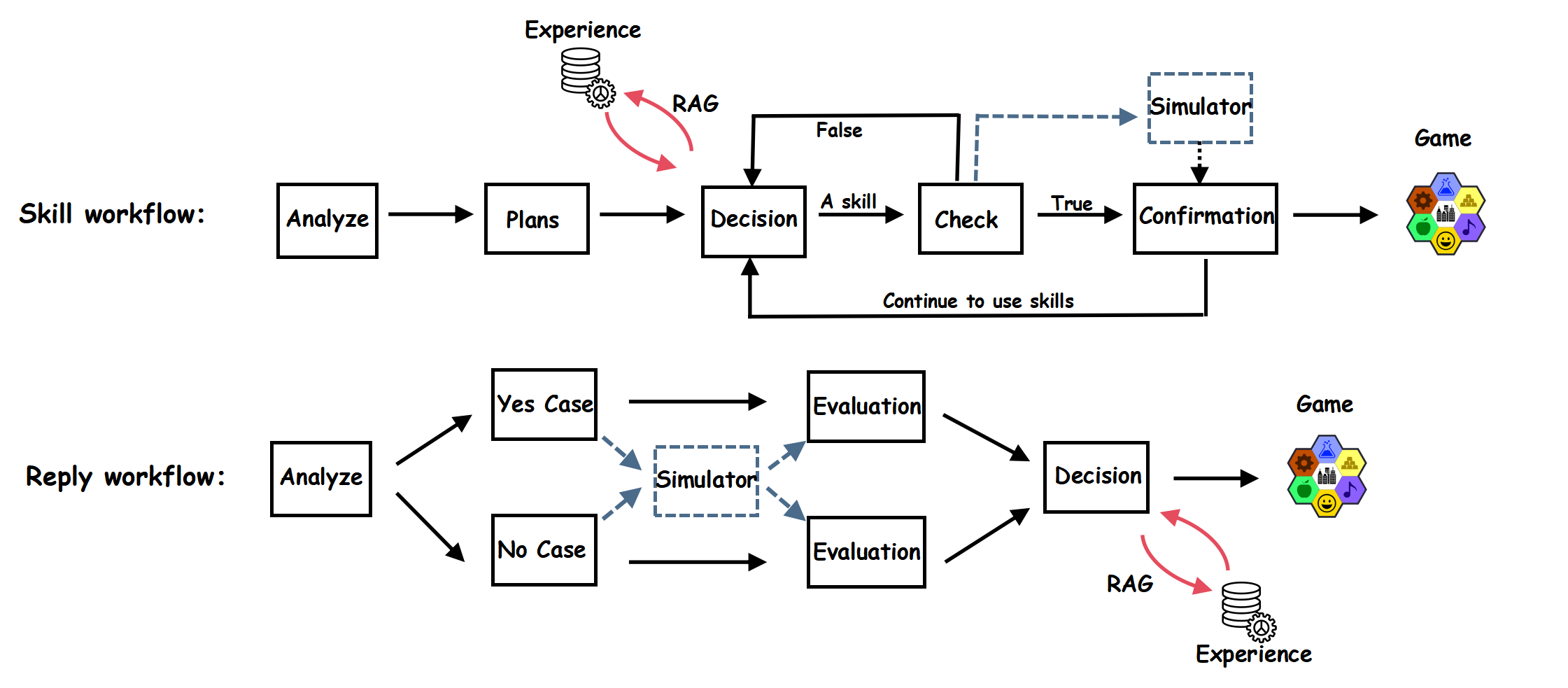}
  \caption{Diagrams for the ``proposing skill'' workflow and ``skill responses'' workflow.\label{Fig:workflow}}
\end{figure*}

In this subsection, we describe how each module is designed and utilized within the entire decision workflow. As depicted in Figure \ref{Fig:action}, each CivAgent not only engages in dialogue during each turn but also proposes skills and responds to the skills suggested by others.  These actions correspond to two distinct workflows within the decision-making process: ``proposing skills'' and ``skills response'',as illustrated in Figure \ref{Fig:workflow}. Each workflow offers two modes of operation; one is standalone, while the other incorporates a simulator to augment numerical decision-making capabilities. The following provides a detailed description of the two workflows:

\begin{itemize}[leftmargin=0pt]
\item\textbf{Workflow for proposing skills.} The workflow consists of the following steps: 1) This process begins with an in-depth reasoning of the current game situation; 2) Based on this reasoning, we establish both long-term and short-term goals aligned with the situation, drawing guidance from historical reflections to inform decision-making; 3) Based on intermediate results from steps 1 and 2, CivAgent derives the optimal decision within the given skills space. we then conduct legitimacy checks on the decisions to ensure that the selected skill comply with the game rules and current environment requirements. If a simulator is integrated, we optimize and adjust the skills based on the feedback provided by the simulator, aiming to achieve the best possible outcome; 4) After repeating step 3 to generate multiple skills, the chosen skills are confirmed and then transformed into actions that comply with the required action interface format.
\item\textbf{Workflow for skill responses.} 
The workflow consists of the following steps: 1) Based on existing game context, detailed skill information, and information of the proponent, CivAgent engages in deep reasoning by employing heuristic questions such as "Why would the opponent propose this?"; 2) Case-by-case discussion is conducted. CivAgent evaluates both acceptance and rejection scenarios for the proposal separately. Here we employ another evaluation LLM to act as a evaluator. If a simulator is used, evaluation can be directly performed through simulation; 3) Finally, intermediate results from steps 1 and 2 are integrated into a prompt, and with Retrieval-Augmented Generation (RAG), historical experiences from similar scenarios are recalled to enable CivAgent to make the most appropriate response decisions.
\end{itemize}
Overall, the design of these two workflows aims to reduce system complexity, decrease reliance on long-context prompt engineering, and improve the interpretability and robustness of the entire decision process. Through this step-by-step strategy, we can more effectively harness the capabilities of large language models, enabling them to operate at their maximum potential within confined problem spaces, thereby enhancing the quality and efficiency of the entire decision-making process.

\subsection{Retrieval Augmented Generation in CivAgent}
We have utilized Llamlndex to construct RAG (Retrieval Augmented Generation) for storing and retrieving reflective experiences of CivAgents accumulated during historical gameplay. after the reflection every several rounds,  a basic fixed-length chunking method is employed to chunk the reflective outcomes. These chunks are subsequently transformed into 4096-dimensional embedding vectors and stored within experience database. In our workflow, situated prior to the decision-making module, we utilize the output of the preceding process as keywords to retrieve the top 5 most similar experiences in the database, leveraging cosine similarity measures. And temporal factors are temporarily not considered.
Here is an example: in a particular game, the long-term and short-term plans for Rome are articulated as follows: ``Our long-term plan is to focus on advancing our scientific research and technology to surpass the Aztecs. Then, we will develop our military to conquer the Aztecs.'' and ``In the short term, we need to efficiently boost our technological levels, seeking research agreements with technologically advanced civilizations such as the Greeks.''. Based on these plans, experiences retrieved from the knowledge base include 1) ``I successfully led Rome to victory through diplomatic cunning and military prowess.  Despite being outmatched in military force initially, But by building strategic alliances and strengthening science and technology, and actively co-existing with technologically advanced civilizations, I eventually turned the tide and established leadership.''; 2) ``Despite facing strong opponents, Rome emerged as the dominant civilization with a formidable force.  Our diplomatic and military strategies were crucial in securing victory, showcasing the strength of our historical and cultural knowledge. The key to our success is to actively sign science and technology agreements and other treaties with other civilizations, maintain development and avoid premature entry into war.'' These experiences are subsequently employed to aid CivAgent in decision-making, enhancing its decision quality in similar scenarios.
\begin{algorithm}[t]
\caption{Reflection by rearview mirror\label{algo1}}
  Initialization Actor, Evaluator, Self-Reflection:\\
   $M_a$, $M_e$, $M_{sr}$\\
  \textbf{Begin Game:}\\
  \text { Generate policy } $\pi_\theta(a_i \mid s_i)$, $\theta=\{M_a, \text { mem }\}$\\
  \text { Generate } $\tau = [a_0, \ldots a_i]$ \text { using }  $\pi_\theta$\\
  \text { Divide } $\tau$ \  into\ segments $\left[\left[a_0,\ldots a_p\right],\left[a_{p+1},\ldots a_j\right],\ldots \left[a_{k+1},\ldots a_i\right]\right]$\\
  Each of the key points to add game feedback, segments $\left[\left[a_0,\ldots a_p,o_p\right],\left[a_{p+1},\ldots a_j,o_j\right],\ldots \left[a_{k+1},\ldots a_i,o_i\right]\right]$\\
    Get the whole sentence game result $R$\\
  \For{each segments}{
  Evaluate segments and $R$ using $M_e$\\
   Generate self-reflection $r$ using $M_{sr}$\\
   Append $r$ \text { to } mem\\
  }
  \textbf{End Game}\\
\end{algorithm}

\begin{algorithm}[!t]
\caption{Reflections with simulator\label{algo2}}
  Initialization Actor,Evaluator,Self-Reflection:\\
   $M_{a}$, $M_{e}$, $M_{sr}$\\
  \textbf{Begin Game:}\\
  \text { Generate policy } $\pi_\theta(a_i \mid s_i)$, $\theta=\{M_a, \text { mem }\}$\\
  \text { Generate } $\tau = [a_0, \ldots a_i]$ \text { using }  $\pi_\theta$\\
  \text { Extract } $\tau$ \  as\ segments $\left[\left[a_p\right],\left[a_{j}\right],\ldots \left[a_{k}\right]\right]$\\
  Simulator feedback is added at each key point, segments $\left[\left[a_p,o_p\right],\left[a_j,o_j\right],\ldots \left[a_k,o_k\right]\right]$\\
  Get the whole sentence game result $R$\\
  \For{each segments}{
  Evaluate segments and $R$ using $M_e$\\
   Generate self-reflection $r$ using $M_{sr}$\\
   Append $r$ \text { to } mem\\
  }
  \textbf{End Game}\\
\end{algorithm}

\subsection{Reflection Mechanism}
\label{reflection}
Traditional reflection frameworks require that an agent can get feedback from the environment immediately after making a decision, and evaluate the score of the decision accordingly.  However, in this game scenario, CivAgent often does not get feedback from the environment immediately after taking diplomatic actions, which makes the traditional reflective approach no longer applicable.
To address this problem,CivAgent introduces two new reflection mechanisms, retrospective learning and simulator-based reflection, to fit the CivSim environment.

\begin{itemize}[leftmargin=0pt]
    \item\textbf{Reflection by rearview mirror:} The core of the rearview mirror reflection lies in the post-game retrospective analysis.  We divided game logs into paragraphs based on keywords (key actions such as "declare war").  The final environmental changes of each segment are then traced back to the corresponding round, combined with past decisions, and fed into the evaluation process.  This allows us to compute a score for each decision, reflect on it, and store the result in our long-term memory.
    \item\textbf{Reflection with simulator:} Simulator-based reflection integrates the counterfactual predictive ability of a simulator to address the limitation of rearview mirror reflection, which can only reflect on a sequence of decisions within a specific timeframe. First, significant decision events are extracted from the game log using specific rules. Subsequently, the game's context at that time is reconstructed, and a simulator is employed to simulate the future impacts of different decisions to aid the reflection process.
\end{itemize}

Algorithm~\ref{algo1} and Algorithm~\ref{algo2} provide pseudocode descriptions for these two reflection mechanisms. Through reflection, CivAgent possesses learning capabilities, enabling it to continuously summarize experiences from environmental feedback. It can swiftly enhance its reasoning and planning abilities by retrieving insights from historical experiences.

\begin{table}[!t]
\centering
\caption{The game situation description for the case study.}
\label{tab:6}
\small
{ 

        \begin{tabular}{cccccc} 
            \toprule 
            Civilization & Civilization score & Technology score & Military score& At war & At research cooperation \\
            \midrule
            Rome&1027 &28 &4733 &N/A &N/A\\
            Greece& 1142& 34 &5263& Egypt& N/A\\
            Egypt&1009 &37 &5180& Greece&N/A\\
            Aztecs& 937& 27& 4051& N/A&N/A\\
            \bottomrule
        \end{tabular}
    }
\end{table}

\section{Experiment}
\subsection{Case Study}
In the following example, we demonstrate the progress of the game by tracing the behavior of the CivAgent Rome during a turn, where the game situation is depicted in Table \ref{tab:6}.

At the beginning of the turn, Rome needs to address various requests from other civilizations. At this moment, Rome notices a research agreement proposal from Egypt. Based on existing intelligence, Rome begins to contemplate: Egypt possesses highly advanced technological capabilities, while Rome's technology level is only moderate. Why would Egypt seek a research agreement with us? Meanwhile, Rome observes that Egypt is currently at war with Greece. Could it be that Egypt aims to use the research agreement as a means to invite Rome to join in the fight against Greece? Reflecting on history, Rome realizes that a hasty acceptance could potentially entangle itself in an unnecessary war. Despite the allure of the research agreement, after careful consideration, Rome decides to reject the request.

At the same time, Rome received a friendly request from Greece. Greece wanted to establish friendly relations with Rome and promised to help it in the future. Rome believed that this might have been a Greek ploy to prevent Rome from aiding Egypt. Now that Egypt had been rejected, it was not in Rome's interest to accept Greece's request, and Rome readily agreed.

At the end of the round, Rome needs to decide which skills to use for this round. By looking at the current situation of the game, Rome made a long-term and short-term plan, and decided that with Egypt and Greece at war, the Aztecs were isolated, and Rome's military strength was far superior to the Aztecs, so it was the best time to attack the Aztecs in order to achieve victory. Rome decided to make a friendly appeal to Egypt, in case Egypt had any doubts about its rejection of the scientific research agreement in this round. At the same time, Rome accepted Greece's friendship request, hoping to deepen the trust between the two sides and reduce the Greek intervention in the attack on the Aztecs. Rome planned to reveal to Greece that Egypt had tried to enlist Rome against the Aztecs, but Rome firmly rejected the offer. So, in this round, Rome declared war on the Aztecs, made a friendly appeal to Egypt, and played a deceitful game to Greece.

After this turn, the game changed dramatically. Greece believed what Rome said, took sides with Rome and offered military and technological help, while the war between Greece and Egypt continued to be fierce, and the Aztecs were attacked by Rome without any help and soon fell. Rome's military, civilization, science and technology and other forces have been rapidly improved, laying the foundation for its victory.

\end{document}